\documentclass[conference]{IEEEtran}
\IEEEoverridecommandlockouts
\usepackage{cite}
\usepackage{amsmath,amssymb,amsfonts}
\usepackage{algorithmic}
\usepackage{graphicx}
\usepackage{textcomp}
\usepackage{xcolor}
\usepackage{balance}
\def\BibTeX{{\rm B\kern-.05em{\sc i\kern-.025em b}\kern-.08em
    T\kern-.1667em\lower.7ex\hbox{E}\kern-.125emX}}
 
\begin{document}

\title{GestLLM: Advanced Hand Gesture Interpretation via Large Language Models for Human-Robot Interaction \\
{\footnotesize \textsuperscript{}}
\thanks{}
}
\author{
\IEEEauthorblockN{Oleg Kobzarev}
\IEEEauthorblockA{\textit{Skolkovo Institute} \\ \textit{of Science and Technology} \\
oleg.kobzarev@skoltech.ru}
\and
\IEEEauthorblockN{Artem Lykov}
\IEEEauthorblockA{\textit{Skolkovo Institute} \\ \textit{of Science and Technology} \\
artem.lykov@skoltech.ru}
\and
\IEEEauthorblockN{Dzmitry Tsetserukou}
\IEEEauthorblockA{\textit{Skolkovo Institute} \\ \textit{of Science and Technology}  \\
d.tsetserukou@skoltech.ru}}
\maketitle

\begin{abstract}
This paper introduces GestLLM, an advanced system for human-robot interaction that enables intuitive robot control through hand gestures. Unlike conventional systems, which rely on a limited set of predefined gestures, GestLLM leverages large language models and feature extraction via MediaPipe\cite{mediapipe_paper} to interpret a diverse range of gestures. This integration addresses key limitations in existing systems, such as restricted gesture flexibility and the inability to recognize complex or unconventional gestures commonly used in human communication.

By combining state-of-the-art feature extraction and language model capabilities, GestLLM achieves performance comparable to leading vision-language models while supporting gestures underrepresented in traditional datasets. For example, this includes gestures from popular culture, such as the ``Vulcan salute" from Star Trek, without any additional pretraining, prompt engineering, etc. This flexibility enhances the naturalness and inclusivity of robot control, making interactions more intuitive and user-friendly.

GestLLM provides a significant step forward in gesture-based interaction, enabling robots to understand and respond to a wide variety of hand gestures effectively. This paper outlines its design, implementation, and evaluation, demonstrating its potential applications in advanced human-robot collaboration, assistive robotics, and interactive entertainment.

\end{abstract}


\maketitle

\section{Introduction}
Gesture-based systems are a promising yet underdeveloped component of human-robot and human-computer interaction\cite{gest_recognition_review}. Current state-of-the-art systems often fall short in delivering natural and intuitive communication. For example, Apple Glass control primarily depends on a limited set of predefined gestures, limiting its capacity to manage complex or culturally nuanced interactions. Some other systems, like from Gestalt Robotics, are concentrated around repeating hand gestures, adapting it to robotic hand movements, or similar robotic components\cite{omnicharger}. While these systems demonstrate the potential of gesture recognition, their inability to use a diverse range of human gestures as potential commands restricts their applications and user experience\cite{llm_for_gesture_selection}.

\begin{figure}[h]
  \centering
  \includegraphics[width=\linewidth]{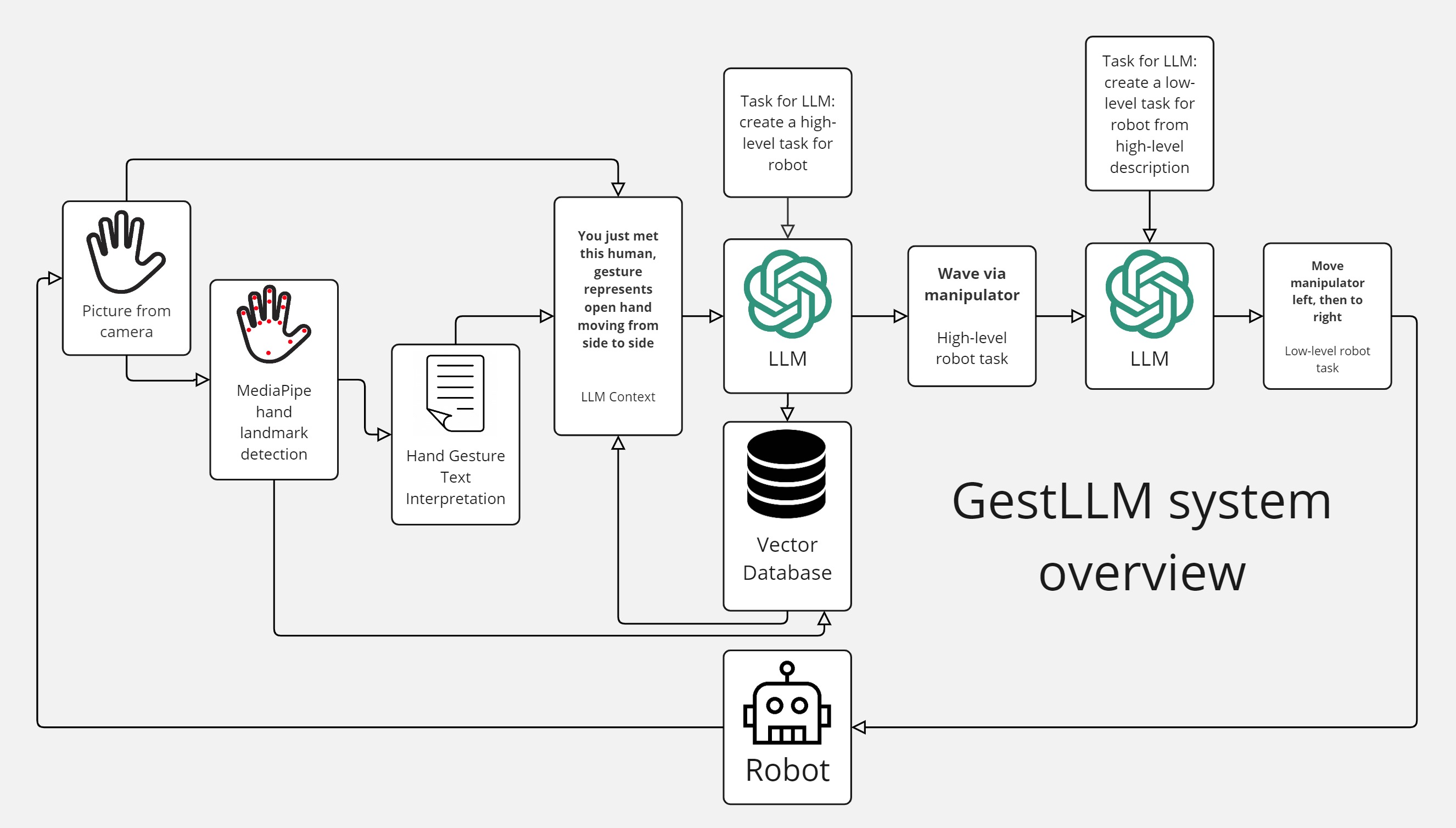}
  \caption{GestLLM system overview.}
  \label{fig:arch}
\end{figure}
Recent advancements in large language models (LLMs), such as GPT-4\cite{openai2023gpt4} and O1\cite{openai_o1}, have demonstrated remarkable performance in tasks like natural language processing, contextual reasoning, and decision-making, and their capabilities have extended into robotics. Building on these advancements, this work introduces a next-generation human-robot interaction system, GestLLM, which utilizes the power of large language models to enable robots to understand complex hand gestures, their context, and more nuances. Unlike traditional gesture-based systems, GestLLM can interpret a diverse range of human gestures that arise naturally in communication. Our goal is to create an advanced and seamless interaction system, incorporating contextual awareness, cultural sensitivity, and adaptability, pushing the boundaries of what gesture-based robotics can achieve. Another researches shows effectiveness of this idea as PC/VR control method\cite{gesturegpt}.

\section{Related Works}
This section reviews works highly aligned with the research focus of this paper or associated with key components and technologies used in the GestLLM system. These works highlight advancements in gesture-based interaction, large language models, and multimodal systems that influence the development of GestLLM.

\textbf{Gesture-based Human-Robot Interaction and Control}: The earliest attempts at hand gesture-based human-robot interaction demonstrated both the potential and challenges of this approach.

\textbf{Gesture-based Human-Robot Interaction for Field Programmable Autonomous Underwater Robots}: Gesture recognition has proven particularly valuable in specialized applications, such as controlling autonomous underwater robots. In scenarios where traditional control devices are impractical due to movement restrictions or durability concerns, gestures provide a feasible and intuitive alternative for human-robot interaction.

\textbf{OmniRace: 6D Hand Pose Estimation for Intuitive Guidance of Racing Drone\cite{omnirace}}: Gesture-based control methods also has proven its effectiveness for drone control. It allows to replace traditional control devices to methods that do not require any additional control hardware and wireless data transmission.

\textbf{Understanding Large-Language Model (LLM)-powered Human-Robot Interaction}:
The integration of large language models (LLMs) into human-robot interaction has demonstrated significant potential for improving communication. By leveraging their ability to interpret complex instructions and context, LLMs enable robots to better understand human intentions, enhancing both the flexibility and efficiency of interactions.

\textbf{CognitiveDog: Large Multimodal Model Based System to Translate Vision and Language into Action of Quadruped Robot}:
This work combines large language models with visual question answering systems to translate high-level, human-understandable instructions into actionable commands for robots, which perfectly integrates with GestLLM ideas.

\section{Approach Overview}

\subsection{Picture preprocessing} 

The initial and crucial step in such systems is picture preprocessing. Various approaches can be utilized to extract more meaningful information from images but it must fit low computational requirements. Low computational requirements on this step are really important to create a responsive interface. After this step we can filter some frames out and not use every frame in the whole pipeline to reduce GestLLM computational requirements without reducing its performance.

Main part of picture preprocessing is the MediaPipe framework and its hand landmark estimation which used as a simple and useful representation of the hand. After MediaPipe preprocessing result is refined with filtering out noise, checking model confidence and including refined results to special structure, related to operator’s hands, additional internal states of hands and more information.

\subsection{Context enhancement}
Context enhancement is a step which makes this system viable on current level of large language models. Current models can't effectively handle information such as sets of coordinates for objects on given scene without pretraining, but pretraining also can't guarantee any level of performance on tasks with such input. Raw picture input also is not effective on gesture recognition tasks because current multimodal large language models struggle with specific image features and specific body part recognition (especially opensource models, such as LLaVA\cite{llava_paper} or MiniGPT-4\cite{zhu2023minigpt}), for example description of each individual finger and it relative position in hand gesture on picture is almost impossible task for most large language models.

Providing adequate context enables large language models to identify gestures accurately, even without direct image input. Technique which gave the best results is specific features extraction from MediaPipe representation and converting such features to their text representation. Currently, best features extracted from keyframes are: individual finger position (in/out of fist), finger direction (angle + direction text description out of 8 options), finger groups (fingers are in group if they are close to each other). Also added text description for hand trajectory between keyframes.

\subsection{Task creation and post-processing}
Large language models allow to convert image + feature text representation input to the task for robots in text form. Text form of task could be classified as one of tasks from a predefined set or as a more complex task for further interpretation. Several researches show how tasks in text form could be transformed to more machine-understandable exact commands for robots: \cite{cognitivedog_paper}, \cite{cognitiveos_paper}. Direct prompts like “find out what task could be assigned to such a gesture” are not really robust and sometimes give strange unexplainable responses (will provide experiment and table for it). Best performance is achieved when previously solved some subtasks: provide name of gesture, provide meaning of gesture. 

Second part is post-processing of the result. For enhancing future results, the current input-output pair is vectorized and added to the vector database. For next recognition this pair could be added to the context of LLM task creator or be used as is instead of using a common computationally complex pipeline. For creation of machine-understandable commands there are two possible steps: classifier and explainer. Classifier checks for existence of currently supported task in output of previous step and in case of existence corresponding command is sent for execution, otherwise (in case of ``explainer") task is divided to such commands (or rejected if division and/or execution are impossible).

\section{Evaluation}

\subsection{Zero-Shot Gesture Recognition Performance}
To evaluate the zero-shot\cite{zero_shot} gesture recognition capabilities of GestLLM, we conducted a set of experiments using gestures that are not commonly included in standard hand gesture datasets. Specifically, we tested recognition accuracy for the ``Vulcan salute" (from the ``Star Trek" science fiction media franchise), the ``shaka sign", the ``finger gun" gesture, and the ``sign of the horns". These gestures are underrepresented or not represented even in hand gestures datasets with wide variety of gestures, such as HaGRID\cite{hagrid}. We compared the performance of GestLLM with GPT-4o\cite{openai2024gpt4ocard} Vision-Language Model (GPT-4o) under the same conditions. Notably, neither GestLLM nor GPT-4o received prior training or prompts for these gestures.

For this task, we collected a custom dataset of approximately 200 images per gesture class. The images were captured in a variety of natural lighting conditions, average hand-to-camera distance is 0.7 m. Images for GestLLM were preprocessed by cropping and scaling them to a resolution of 640x480 pixels, while the original images with a resolution of 1280x720 pixels were used directly for GPT-4o.

\subsubsection{Recognition Accuracy at Close Range}

Recognition accuracy was first measured at a close range of approximately 0.7 meters from the camera. Table~\ref{tab:accuracy_close} summarizes the results.

\begin{table}[h]
    \centering
    \caption{Recognition Accuracy at Close Range (0.7 m)}
    \label{tab:accuracy_close}
    \begin{tabular}{|c|c|c|}
        \hline
        Gesture & GestLLM (\%) & GPT-4o (\%) \\
        \hline
        Vulcan salute & 94.04 & 94.72 \\
        Shaka sign & 89.71 & 90.38 \\
        Finger gun & 93.40 & 92.53 \\
        Sign of the horns & 95.73 & 91.75 \\
        \hline
    \end{tabular}
\end{table}

At close range, GestLLM achieved accuracy levels comparable to GPT-4o, with slight differences depending on the gesture. For instance, GestLLM outperformed GPT-4o on the ``sign of the horns" gesture, achieving a recognition accuracy of 95.73\% compared to 91.75\%. Conversely, GPT-4o showed marginally better results for the ``shaka" sign. Overall, both methods demonstrated strong performance at this distance, but computational requirements for GestLLM are much lower than GPT-4o.

Also, both systems were evaluated at longer ranges (2-4 m). While GestLLM maintained stable recognition accuracy even at increased distances until distance reached limit of MediaPipe recognition capabilities (about 4 m). Performance of GPT-4o dropped significantly after distance reached 2-2.5 m on our setup and was pretty unstable, with accuracy ranging between 20\% and 50\% depending on the gesture and distance. This sharp decline can be attributed to limitations in GPT-4o's vision capabilities, particularly for gestures requiring precise finger positioning at greater distances. In contrast, GestLLM demonstrated robust performance, maintaining similar accuracy levels until MediaPipe, the underlying hand-tracking framework, began failing to detect detailed finger positions at approximately 4 meters. These results could be improved by changing preprocessing framework to something more advanced than MediaPipe, for example OpenPose's hand detection component\cite{simon2017hand}, or trying more image preprocessing steps\cite{preprocessing_review}.

\subsubsection{Results}
The results highlight the strong zero-shot recognition capabilities of GestLLM, which matches the performance of GPT-4o at close distances and significantly outperforms it at greater ranges. GestLLM's ability to maintain high accuracy across varying distances demonstrates its robustness in interpreting complex and uncommon gestures,
\begin{figure}[h]
  \centering
  \includegraphics[width=\linewidth]{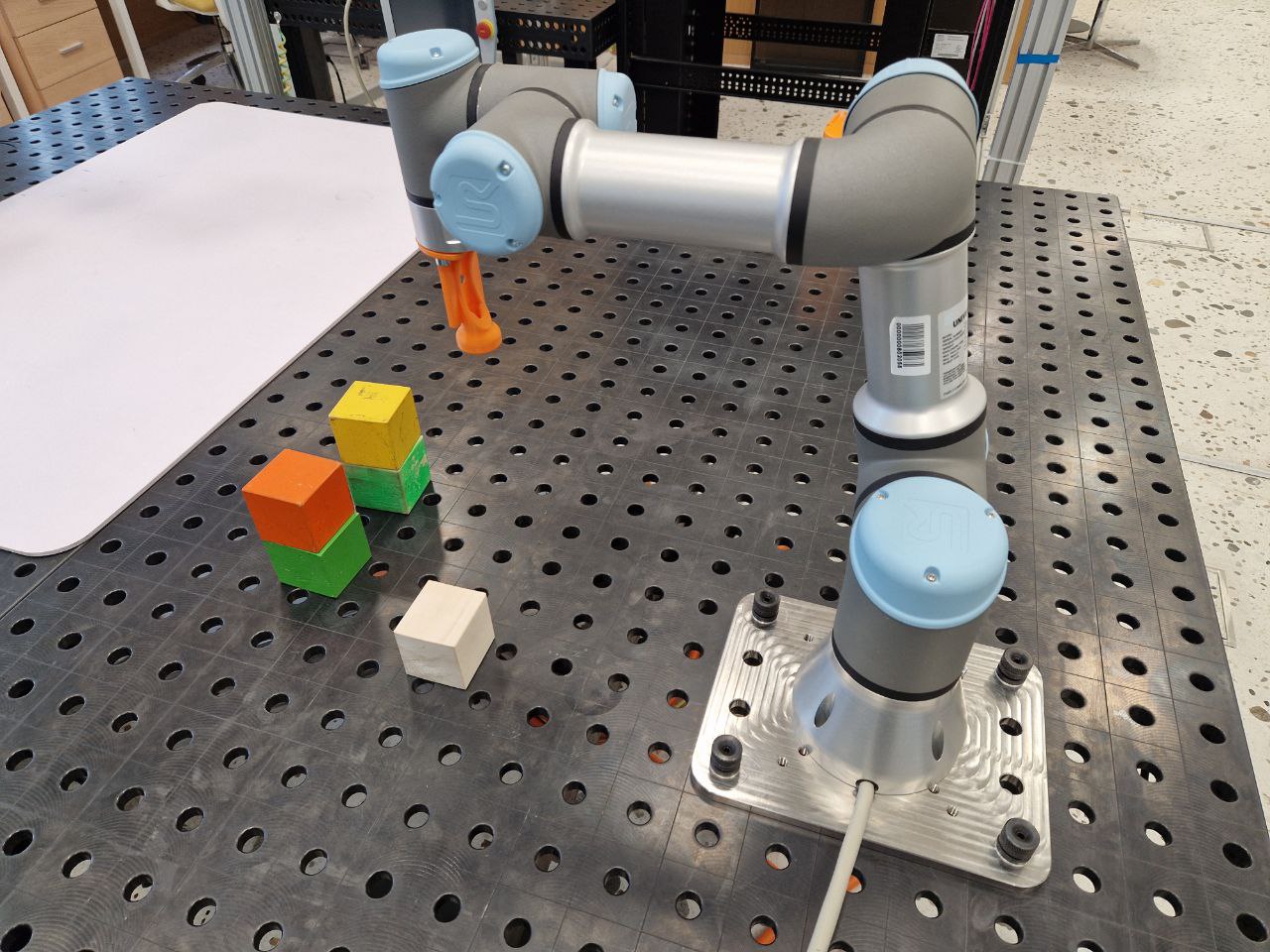}
  \caption{Universal Robots UR3 Manipulator.}
  \label{fig:arch}
\end{figure}
\newline even in challenging scenarios. These findings suggest that GestLLM provides a reliable and scalable solution for gesture-based interaction in real-world applications where distance and gesture variability are key factors.

\subsection{Robot control}
To evaluate the effectiveness and usability of the proposed GestLLM system, we conducted a comparative study using a robotic hand as the interaction platform. The study compared two control methods: (1) the GestLLM-based hand gesture control system, and (2) a gamepad control method. Both systems were used to perform a series of predefined tasks involving the manipulation and control of a robotic manipulator.

The experimental setup utilized the Universal Robots UR3 manipulator. To ensure safety and consistency across both control methods, the speed, acceleration, and operating area of the manipulator were restricted and standardized for all experiments.

The tasks were specifically designed to evaluate the capabilities of the GestLLM system and its integration with hand tracking for controlling a real robotic hand. Each participant completed the tasks using both control methods, with the order of methods randomized. Prior to beginning the tasks, participants underwent a brief training session to familiarize themselves with both control methods. For evaluation was used NASA Task Load Index\cite{nasatlx_paper} (NASA TLX) method. Participants provided written informed consent before experiments. The consent form included information about the purpose, procedures and the voluntary nature of participation.

The task sets were divided into the following categories:
\begin{itemize}
  \item Basic Movement Tasks: These tasks focused on movement control using either hand tracking (for GestLLM) or joysticks (for the gamepad):
    \begin{itemize}
      \item Drawing figures in the air (e.g., circles, lines)
      \item Pushing objects (color-coded cubes) with the end effector of the manipulator
    \end{itemize}
  \item Movement with Program Activation: These tasks combined movement control with the activation of specific programs at designated times and positions, using gestures (for GestLLM) or buttons (for the gamepad):
    \begin{itemize}
      \item Moving within a single plane to target and activate a program for pushing objects outside of the reachable area in that plane
      \item Drawing figures in the air with action triggers at specific, pre-defined points (unknown to participants, told by assistant in process).
    \end{itemize}
\end{itemize}

\subsubsection{Results}
The results of the NASA TLX evaluation are summarized in Table~\ref{tab:workload_scores}. Overall, we analyzed differences in workload and performance between the GestLLM and gamepad control methods to assess the benefits of our proposed system.

\begin{table}[h]
    \centering
    \caption{NASA TLX Scores for Each Control Method}
    \label{tab:workload_scores}
    \begin{tabular}{|c|c|c|}
        \hline
        Dimension & GestLLM & Gamepad \\
        \hline
        Mental Demand & 44.3 & 37.1\\
        Physical Demand & 17.4 & 6.8\\
        Temporal Demand & 18.5 & 20.8\\
        Effort & 27.2 & 28.4 \\
        Frustration & 10.3 & 11.0 \\
        Performance & 27.9 & 25.8\\
        \hline
    \end{tabular}
\end{table}

\subsubsection{Analysis and Discussion}
The NASA TLX results summarized in Table~\ref{tab:workload_scores} provide important insights into the usability and workload associated with the GestLLM and gamepad control methods. Key observations are as follows:

\begin{itemize}
    \item \textbf{Mental Demand:} The GestLLM method scored higher on Mental Demand (44.3) compared to the gamepad (37.1). This suggests that participants required more cognitive effort to interpret and utilize the gesture-based system, likely due to the novelty of the interface.
    
    \item \textbf{Physical Demand:} GestLLM exhibited a higher Physical Demand score (17.4) compared to the gamepad (6.8). This increase can be attributed to the physical effort required for precise hand gestures, whereas the gamepad relies on minimal finger movement.
    
    \item \textbf{Temporal Demand:} Temporal Demand scores were comparable between the two methods (18.5 for GestLLM vs. 20.8 for gamepad), indicating that both systems imposed similar time-related pressure on participants during task completion.
    
    \item \textbf{Effort and Frustration:} GestLLM scored slightly lower in Effort (27.2) and Frustration (10.3) compared to the gamepad (28.4 and 11.0, respectively). This reflects the intuitive nature of gesture-based control, which participants found less stressful and cognitively exhausting despite its novelty.
    
    \item \textbf{Performance:} Performance scores were comparable between GestLLM (27.9) and the gamepad (25.8), indicating that both methods enabled participants to accomplish tasks effectively.
\end{itemize}

\noindent These findings highlight the strengths of GestLLM in providing a user-friendly interface while balancing cognitive and physical demands. Despite the slightly higher Mental and Physical Demands, the lower Frustration and Effort scores indicate that participants adapted well to the system and found it intuitive overall.

The results also support the hypothesis that gesture-based control, when combined with advanced language models, can offer a viable alternative to traditional control methods. GestLLM’s ability to handle complex gestures and contextual nuances makes it a promising tool for human-robot interaction. Future research could focus on refining the system to reduce Mental and Physical Demands further, enabling a more seamless user experience.

\section{Conclusion}

In this paper, we introduced GestLLM, an advanced human-robot interaction system that utilizes large language models to interpret and execute complex hand gestures, accounting for contextual and cultural nuances. GestLLM addresses significant limitations of conventional gesture-based systems, such as restricted gesture sets and limited contextual understanding, enabling more natural and intuitive communication between humans and robots.

Our evaluation included two aspects: (1) a comparison with a traditional gamepad control method using the NASA TLX framework and (2) an assessment of zero-shot gesture recognition accuracy for uncommon gestures. The NASA TLX results indicated that GestLLM provides a comparable overall workload experience to the gamepad while excelling in reducing Frustration and Effort. The slightly higher Physical and Mental Demand scores for GestLLM highlight the system’s novelty and the additional effort required for precise hand gestures. Nevertheless, participants rated the system as intuitive and effective, showcasing its potential for natural human-robot interaction.

In zero-shot gesture recognition tests, GestLLM maintained robust performance up to approximately 4 meters, where limitations in MediaPipe-based gesture feature extraction began to affect accuracy. In same conditions, GPT-4o experienced significant accuracy degradation beyond 2–2.5 meters in our setup. This capability reduces the need for users to move closer to the robot, enhancing convenience and practicality, especially in scenarios like remote robotic control, hazardous environments, or large-scale industrial settings where maintaining physical distance is advantageous.

These findings underscore GestLLM’s potential to revolutionize gesture-based human-robot interaction, making it more robust, adaptable, and user-friendly for diverse real-world applications. Future work will aim to reduce Mental and Physical Demands by refining the gesture recognition process and improving training protocols. Additionally, expanding GestLLM's capabilities to include dynamic and multi-step gestures and enhancing its adaptability for diverse real-world applications will be prioritized. 

\section*{Acknowledgements} 
Research reported in this publication was financially supported by the RSF grant No. 24-41-02039.

\vspace{12pt}

\end{document}